\documentclass[runningheads]{llncs}

\usepackage{eccv}

\usepackage{eccvabbrv}

\usepackage{graphicx}
\usepackage{amsmath}
\usepackage{amssymb}
\usepackage{booktabs}
\usepackage{customization}
\usepackage{multirow}
\usepackage[dvipsnames]{xcolor}
\usepackage{booktabs, multirow, array, placeins, pifont, enumitem, bm, soul, nicefrac, bbold, boldline, tocloft, wrapfig, graphicx, caption, comment, xspace, colortbl}

\usepackage[accsupp]{axessibility}

\newcommand{\red}[1]{\textcolor{red}{#1}}
\newcommand{\green}[1]{\textcolor{green}{#1}}
\newcommand{\pink}[1]{\textcolor{magenta}{#1}}
\newcommand{\cyan}[1]{\textcolor{cyan}{#1}}

\newcommand{\method}[1]{{\textbf{CLIX\textsuperscript{3D}}}}

\usepackage{customization}

\definecolor{checkmark}{HTML}{40826D}
\definecolor{xmark}{HTML}{E62020}

\usepackage{arydshln}
\makeatletter
\def\adl@drawiv#1#2#3{%
        \hskip.5\tabcolsep
        \xleaders#3{#2.5\@tempdimb #1{1}#2.5\@tempdimb}%
                #2\z@ plus1fil minus1fil\relax
        \hskip.5\tabcolsep}
\newcommand{\cdashlinelr}[1]{%
  \noalign{\vskip\aboverulesep
           \global\let\@dashdrawstore\adl@draw
           \global\let\adl@draw\adl@drawiv}
  \cdashline{#1}
  \noalign{\global\let\adl@draw\@dashdrawstore
           \vskip\belowrulesep}}
\makeatother

\usepackage[pagebackref,breaklinks,colorlinks,citecolor=eccvblue]{hyperref}

\usepackage{orcidlink}

\newcommand\blfootnote[1]{%
  \begingroup
  \renewcommand\thefootnote{}\footnote{#1}%
  \addtocounter{footnote}{-1}%
  \endgroup
}

\begin{document}

\title{Multimodal 3D Object Detection on Unseen Domains}

\titlerunning{\method~}

\author{Deepti Hegde\inst{1}$^\star$\and
Suhas Lohit\inst{2} \and 
Kuan-Chuan Peng\inst{2} \and \\
Michael J. Jones\inst{2} \and
Vishal M. Patel\inst{1} }

\authorrunning{Hegde et al.}

\institute{Johns Hopkins University
\\
 \and
Mitsubishi Electric Research Laboratories (MERL)\
}

\maketitle

\begin{figure}
    \centering
    \includegraphics[width=\linewidth]{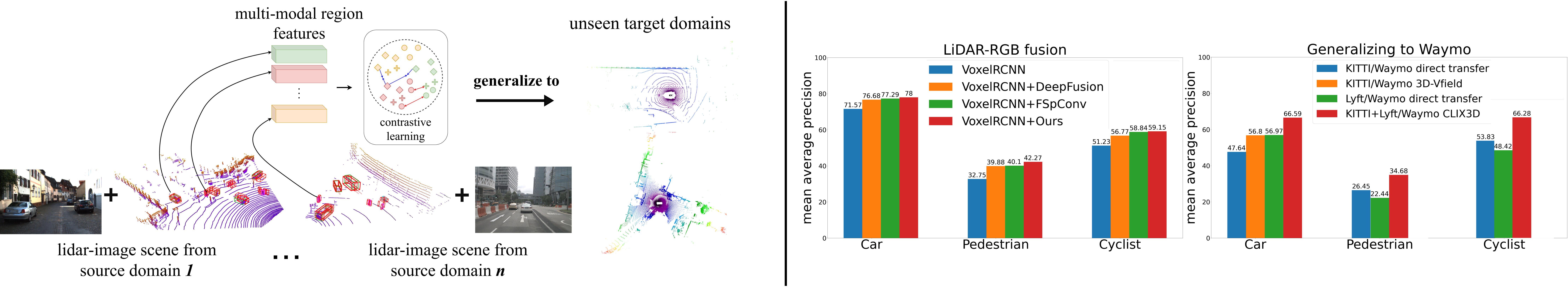}
    \caption{(Left) Overview of the contrastive learning framework of \method~. We use multi-source training and supervised contrastive learning between region level features to improve robustness of 3D LiDAR and LiDAR-image based object detection networks. (Middle-left) Comparison of 3D detection precision performance of \method~ against LiDAR-only detection and SOTA fusion methods. (Far-ight) Comparison of \method~'s robustness to unseen distributions with SOTA domain generalization work.} 
    \vspace{-0.2in}
    \label{fig:teaser}
\end{figure}

\vspace{-0.2in}

\blfootnote{$^\star$ This work was done when Deepti Hegde was an intern at MERL.}

\begin{abstract}
 LiDAR datasets for autonomous driving exhibit biases in properties such as point cloud density, range, and object dimensions. As a result, object detection networks trained and evaluated in different environments often experience performance degradation. Domain adaptation approaches assume access to unannotated samples from the test distribution to address this problem. However, in the real world, the exact conditions of deployment and access to samples representative of the test dataset may be unavailable while training. We argue that the more realistic and challenging formulation is to require robustness in performance to \emph{unseen} target domains. We propose to address this problem in a two-pronged manner. First, we leverage paired LiDAR-image data present in most autonomous driving datasets to perform multimodal object detection. We suggest that working with multimodal features by leveraging both images and LiDAR point clouds for scene understanding tasks results in object detectors more robust to unseen domain shifts. Second, we train a 3D object detector to learn multimodal object features across different distributions and promote feature invariance across these source domains to improve generalizability to unseen target domains. To this end, we propose \method~, a multimodal fusion and supervised contrastive learning framework for 3D object detection that performs alignment of object features from same-class samples of different domains while pushing the features from different classes apart. We show that \method~yields state-of-the-art domain generalization performance under multiple dataset shifts.
\end{abstract}

\section{Introduction}
\label{sec:intro}

LiDAR point clouds provide direct, albeit sparse, 3D geometric information of a scene through accurate depth estimates, while RGB images can provide high resolution 2D color and texture information. Recent years have seen the release of several large scale multimodal datasets containing registered LiDAR point clouds and RGB images \cite{nuscenes2019,waymo,KITTI,chang2019argoverse,lyft}, which have aided the development of numerous deep neural networks for solving perception tasks for autonomous navigation such as segmentation \cite{milioto2019rangenet++,zhu2021cylindrical,aygun20214d} and object detection\cite{yan2018second, shi2019pointrcnn, shi2021parta2,shi2020pv,lang2019pointpillars}. Until recently, the best performing 3D object detectors operate only on LiDAR point clouds \cite{shi2020pv,shi2019pointrcnn,lang2019pointpillars,shi2021parta2}, despite the availability of paired image-point cloud data in most autonomous driving datasets \cite{nuscenes2019,KITTI,waymo,lyft,chang2019argoverse}. The image and LiDAR modalities offer complementary information while describing the same underlying scene, making the multimodal fusion approach a natural choice for training a detection network. Several recent works have demonstrated the effectiveness of LiDAR-image fusion for 3D scene understanding \cite{li2022deepfusion,chen2022focal,liu2022bevfusion,jacobson2023center}, and outperformed LiDAR-only methods. In this work, we contribute to this direction and propose a multi-stage LiDAR-image fusion method for 3D object detection. In particular, we focus on how multimodal object detection networks can be trained to be robust when evaluated on scenes ``in-the-wild'' (see Fig. \ref{fig:teaser}).

LiDAR point clouds collected from different environments vary widely between one another in terms of point cloud density, scene properties and object dimensions due to different modes of capture, locations, weather conditions, \etc. LiDAR scenes from different datasets have large differences that are easily visible. 

\begin{wrapfigure}{r}{0.5\textwidth}
    \centering
    \includegraphics[width=0.48\textwidth]{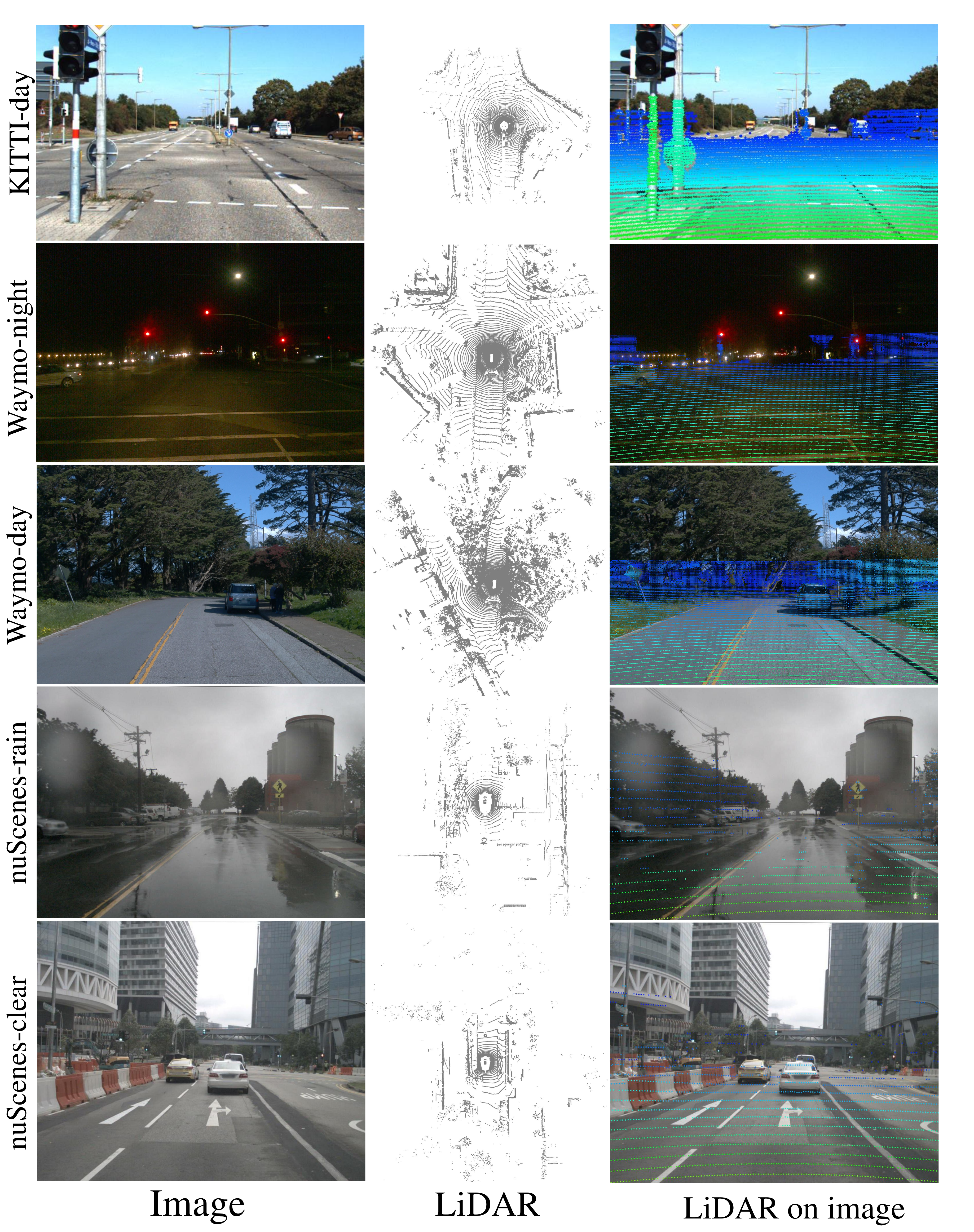}
    \caption{LiDAR and image scene examples from the KITTI \cite{KITTI}, Waymo \cite{waymo} nuScenes \cite{nuscenes2019} datasets in different environmental conditions, which are listed at the beginning of each row. Images are particularly prone to illumination conditions, while LiDAR scenes differ in density between datasets. The first column shows the image scene, the second column is a bird's-eye-view (BEV) of the LiDAR scene and the last row shows the LiDAR scene projected to the front image.}
    \vspace{-0.2in}
    \label{fig:dataset}
\end{wrapfigure}

While the severity of the cross-dataset distribution gap is not as large when dealing with images, changes in the time-of-day and adverse weather during capture can result in datasets that are biased to specific conditions.
This becomes a problem when performing scene understanding tasks in real-world scenarios, where the conditions of capture at test time may differ from the training dataset. The performance of neural network models trained on data from a particular distribution drops when shown samples from a different distribution \cite{Wang2020TrainIG}. 
We suggest that including image information helps not only the baseline performance, but also in training networks robust to distribution shifts. Images provide dense color and texture information, while LiDAR point clouds provide sparse but accurate depth measurements. 
Data from both these modalities is prone to varying highly depending on the environmental conditions. Images are particularly prone to illumination changes, while LiDAR scenes are not. The density and range of point cloud scenes are dependent on the sensor specification, and tend to differ largely between datasets, while the corresponding images are unaffected. This can be observed in Fig. \ref{fig:dataset}, where LiDAR-image pairs from the Waymo Open Dataset \cite{waymo}, nuScenes \cite{nuscenes2019} and KITTI \cite{KITTI} are shown side-by-side to demonstrate the differing properties of each modality in differing conditions.
We suggest that the complementary nature of both the information and sources of domain shift makes LiDAR-RGB fusion the natural choice to improve the robustness of 3D object detection networks to distribution shift. DeepFusion \cite{li2022deepfusion} briefly explores this capability by evaluating their proposed LiDAR-RGB fusion pipeline on location-specific domain shift within the same dataset. In contrast, we address more challenging cross-dataset domain shifts and propose object-level contrastive learning to build more robust detectors.

Most current works that address the problem of domain gaps assume the existence of an unlabeled dataset that is representative of the domain used for evaluation (target domain). These methods further adapt a source network using target samples and at times labelled source samples to perform unsupervised domain adaptation (UDA) \cite{yang2021st3d, Saltori2020SFUDA3DSU,scalablePseudo,xu2021spg,Luo2021multiLevel,st3dpp,hegde2023source,hegde2021attentive}. This formulation can be  unrealistic, as the target domain characteristics are often unknown and can change dynamically. Additionally, the adapted model is suitable only for the target domain it is trained for, and must be re-trained for every new target distribution.
In contrast, we formulate and propose a method to address the domain \textbf{generalization} (DG) problem, which is a more practical and challenging setting for the 3D object detection task \cite{NIPS2011_b571ecea,zhou2022domain}. In the DG setting,  no information about the target domain(s) is available during training. The guiding principle of DG is that training a model able to generalize over diverse source domains can help generalize to unseen target distributions \cite{ben2006analysis}.

Gulrajani and Lopez-Paz~\cite{gulrajani2020search} have shown that performing empirical risk minimization across multiple diverse datasets results in highly generalizable models given enough data and an effective model selection strategy. DG through domain-invariance has also received significant attention for 2D images, especially for image recognition~\cite{yao2022pcl}, including methods based on contrastive learning. However, extending these ideas to 3D object detection is not straightforward due to relative complexity of the scene compared to samples from image classification datasets, as well as the more difficult task of performing denser predictions. We suggest that this problem setting requires operating on more local regions in the scene. We focus on individual objects in a scene by \textbf{utilizing region features provided by existing object detection networks to enforce domain invariance} in the feature extraction backbones of 3D object detectors.
 
We thus propose to tackle performance degradation of object detectors evaluated on unseen domains in a two-pronged manner. First, we propose a multimodal fusion method for RGB and LiDAR data to leverage complementary information across modalities for increased robustness. Second, we design a novel framework called Contrastive LiDAR+Image Cross-Domain 3D Object Detection -- \method~ for  training LiDAR and LiDAR-RGB 3D object detectors to improve performance on unseen domains by learning domain invariant representations of objects in a scene. Fig. \ref{fig:teaser} shows an overview of the contrastive training method. The features from object regions of each scene are aligned in a common embedding space according to category to train the network to learn properties of the object that are invariant to domain specific conditions. We are the first to propose a multi-source multimodal setting to address robustness to unseen domains. \method~ outperforms multi-source training and single-source DG on most domain shift scenarios.

\begin{enumerate}[label=\arabic*., leftmargin=*, topsep=0pt]
\setlength\itemsep{-0.0em}
    
    \item We propose \method~, a supervised contrastive learning framework for learning domain-invariant features that are suited for 3D object detection by operating on small regions rather than at the global level. 
    \item By leveraging various autonomous driving datasets acquired using different sensors under different conditions and environments, we show that \method~ results in improved domain generalization to unseen target domains. 
    \item We propose a new LiDAR-RGB fusion approach -- MSFusion -- as a part of \method~ that improves same-domain and cross-domain 3D object detection performance, compared to state-of-the-art (SOTA) fusion methods.
\end{enumerate}

\section{Related works}
\label{ref:related_work}

\subsection{3D object detection}
There are fundamentally three main approaches to performing purely LiDAR-based object detection with deep networks in literature -- (a) operating directly on the irregular point clouds \cite{yang20203dssd,he2020structure}, (b) first voxelizing or pillarizing the point clouds and operating on the voxels \cite{lang2019pointpillars,second,shi2021parta2,deng2021voxel}, and (c) projecting the point clouds to 2D, \eg, bird's-eye view (BEV) \cite{liu2021bev,huang2022tig}. 
Each network may also be characterized as a single-stage or a two-stage network. Single stage object detectors directly regress and classify bounding box predictions from features whereas two-stage object detectors use an additional refinement head operating on region proposals. The SOTA single stage object detector SE-SSD \cite{zheng2021se} uses IoU-based matching to align a student-teacher network to perform soft filtering. VoxelRCNN \cite{deng2021voxel} proposes a novel region-of-interest pooling and box refinement and outperforms previous point and voxel based methods. In this work, we use VoxelRCNN as the base object detector due to its superior performance as a two-stage object detector. (See Sec. \ref{sec:exp_set})

\noindent\textbf{LiDAR-image fusion:}
Several works seek to incorporate image data to boost 3D object detection performance. Early methods such as PointPainting \cite{vora2020pointpainting} and PointAugmenting \cite{wang2021pointaugmenting} concatenate the input LiDAR point cloud with semantic scores or deep features before being passed into the 3D network. Current single-view, convolutional approaches improve on these by performing fusion at the feature level mid-way through the network. DeepFusion \cite{li2022deepfusion} performs fusion by aligning the LiDAR and image scenes through inverting the LiDAR augmentations before fusing at the feature level using a cross-attention mechanism for feature selection. FocalsConv \cite{chen2022focal} proposes a method for sparse convolutions with learned sparsity which they leverage for LiDAR-image fusion. TransFusion \cite{bai2022transfusion} and LIFT \cite{zeng2022lift} are transformer based fusion approaches. BEVFusion \cite{liu2022bevfusion} leverages multi-view images to create bird's-eye-view image feature maps for fusion. In this work, we focus on single view convolutional fusion approaches due to their compatibility with existing LiDAR detectors and datasets.

\subsection{Cross-domain transfer}
A significant number of works in perception literature address dealing with domain shift, by either adapting to specific targets, or training robust networks.

\noindent\textbf{Domain adaptation for 3D object detection:}
Recent works that address unsuerpvised domain adaptation (UDA) for 3D object detection include ST3D~\cite{yang2021st3d}, ST3D++~\cite{st3dpp}, MLCNet~\cite{Luo2021multiLevel}, and scalable pseudo-labeling~\cite{scalablePseudo}. However, these  works use a \textbf{single} source domain and assume access to annotated or unannotated samples from the target domain that are used for model adaptation. While Wang's method~\cite{Wang2020TrainIG} does not train on target data, it uses the bounding box statistics of the target dataset to resize predicted bounding boxes to perform adaptation. Some works train with multiple source domains \cite{yao2021multi} or models \cite{tsai2023ms3d}, but still operate in the UDA formulation. Zhang \etal \cite{zhang2023uni3d} show that multi-source training can aid in boosting detection performance, but do not perform cross-dataset transfer.

\noindent\textbf{Domain generalization:}
In contrast, in this paper, we study a related but novel setting for 3D object detection -- multi-source DG -- where the target domain is completely unseen during training. We show that multi-source and multimodal training from diverse domains can lead to improved generalizability in 3D object detection models. We design a supervised contrastive learning method \method~ that can be used with LiDAR and LiDAR-RGB detector architectures to lead to higher generalization performance. Lehner \etal~\cite{lehner20223d} recently propose 3D-VField, an adversarial data augmentation strategy, for DG with point clouds. However, they only show single-source DG and do not discuss domain invariance.  Although the idea of contrastive learning has been employed for DG in 2D image recognition~\cite{yao2022pcl}, it is not straightforward to apply such ideas for 3D object detection from LiDAR point clouds and we discuss the challenges in design and implementation in Sec. \ref{sec:method}.

\section{\method~}
\label{sec:method}
We first describe the multi-stage LiDAR-image feature fusion approach, followed by the supervised contrastive learning framework for multi-source object detection.

\subsection{LiDAR-image fusion}
\label{sec:li}
We design a multi-stage fusion module, called \textbf{MSFusion}, for LiDAR and image data that are incorporated into existing 3D object detection networks to consider image information during the feature extraction stage. Consider a LiDAR scene and image pair $\{p,q\}$ such that $p \in \mathbb{R}^{g\times3} $ and $q \in \mathbb{R}^{h\times w \times 3}$, where $g$ denotes the number of points in the LiDAR scene and $h\times w$ denotes the spatial dimensions of the image. Let $\boldsymbol{\phi}$ be a convolution-based encoder for the image modality decomposed into $s$ stages $\{\phi_i\}_{i=1}^{s}$ and $\boldsymbol{\psi}$ be a 3D feature extraction network that processes point clouds decomposed into $\{\psi_i\}_{i=1}^{s}$. 
The image is passed to $\boldsymbol{\phi}$ to obtain a set of feature maps $F_{\phi_i}$ of successively smaller feature dimensions from each stage. The 3D encoder consists of a series of 3D convolution blocks and set abstraction layers that process a voxelized point cloud to give 3D feature maps consisting of stacked voxel features. The voxelized LiDAR scene is passed to $\boldsymbol{\psi}$ to obtain sets of voxels that undergo successive set abstractions to give sets of features with reducing spatial dimensions $F_{\psi_i}$. For each stage pair $\{F_{\phi_i},F_{\psi_i}\}$ we find the voxel feature to pixel feature correspondence before performing deep feature fusion.

\noindent\textbf{Finding voxel-pixel correspondence.}
  Given the camera matrix, each voxel may be projected to the image feature plane to obtain pixel-voxel level correspondences. Since the number of points that represent the scene are fewer than the number of pixels ($m < h\times w$), we assign a neighborhood of pixels of size $k$ to each projected point, with the pixel location of the point as the center of the neighborhood. Each LiDAR point now corresponds to a $[k \times k]$ patch of the image. With reducing spatial dimensions of each 2D feature map, the camera intrinsics are scaled accordingly.

\noindent\textbf{LiDAR augmentation reversal.}
The use of data augmentations are standard practice in 3D LiDAR-only object detection networks, as they aid in performance \cite{shi2021parta2,deng2021voxel,second}. However, these augmentations are unique to the LiDAR modality and include transformations such as rotation and scaling, which affect the correspondence of points with pixels. To ensure that the 3D encoder benefits from these augmentations and that accurate voxel-pixel mapping takes place, we keep a record of each augmentation type and degree of augmentation and perform reversal in the voxel feature space before the fusion step.

\noindent\textbf{Deep feature fusion.}
Consider the set of LiDAR-image feature maps $\{F_{\phi_i},F_{\psi_i}\}_{i=1}^s$. After applying the reverse augmentation on the voxel feature maps and obtaining voxel-pixel correspondences for each stage, each feature vector path is average pooled, concatenated along the feature dimension and passed to a projection layer that embeds the feature into the 3D embedding space. The resultant feature is passed to the next stages of the object detection network.

\subsection{Multi-source 3D object detection}
\label{sec:scl}

\begin{figure*}
    \centering
    \includegraphics[width=\linewidth]{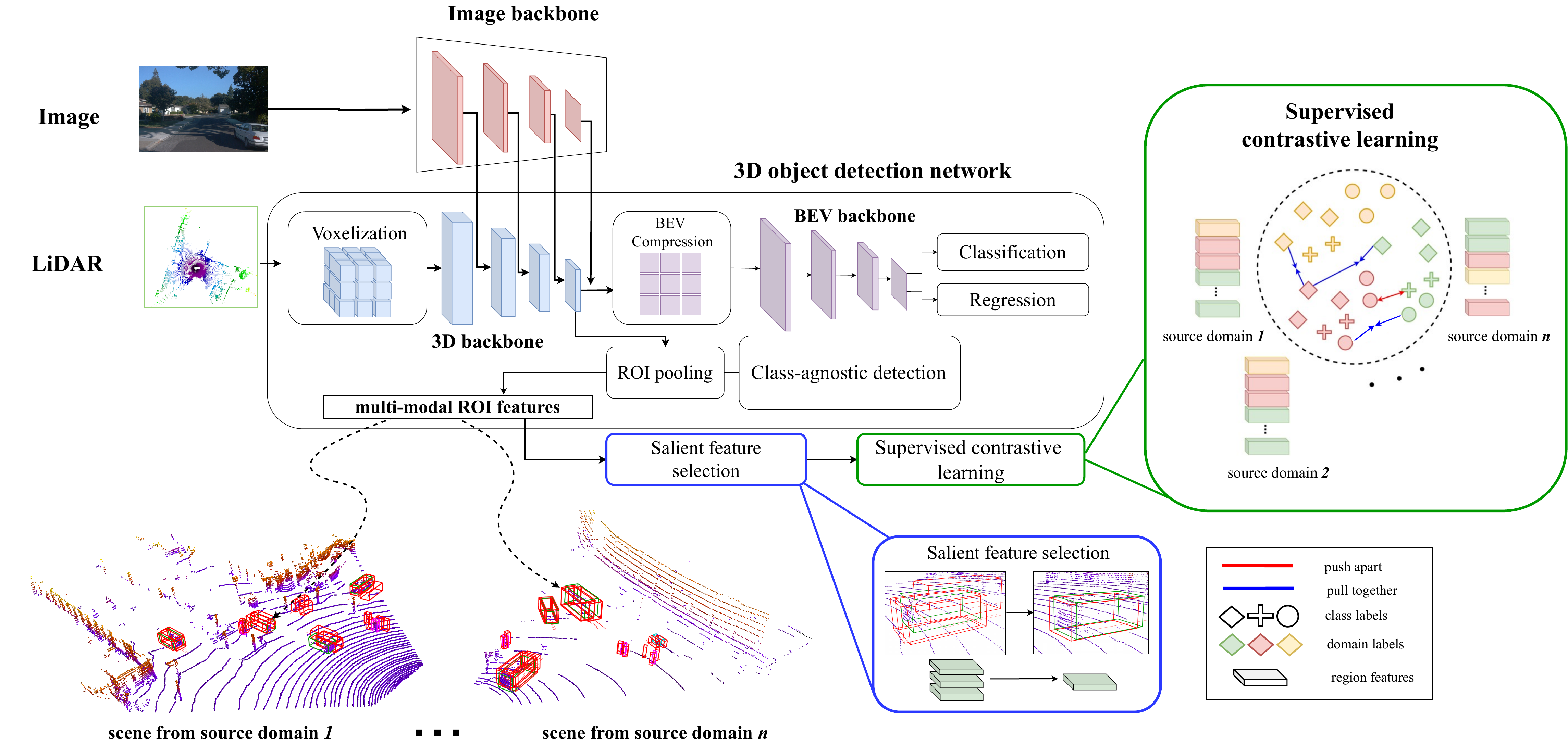}
    \caption{Description of the proposed \method~ for generalizing 3D object detectors to unseen target domains. Samples from multiple diverse source domains are used to train the object detectors. Following multi-stage deep feature fusion in the feature extraction backbone, supervised contrastive learning is applied on ROI features obtained after the pooling step which encourages domain invariance. As illustrated on the right side, region features that belong to the same class, but from different domains are encouraged to be closer together in feature space, while those that belong to different classes are pushed apart.}
    \vspace{-0.15in}
    \label{fig:main}
\end{figure*}

We now focus on the later stages of the object detection network. The methodology for contrastive multi-source training applies to both single and multimodal object detectors. For the purpose of clarity, we consider LiDAR object detectors here. The task of 3D object detection is to localize and classify salient objects in the scene. This is done by estimating location and dimensions of a 3D bounding box that describe the position of each object, and classifying it into one of the predefined classes. Let a point cloud dataset of $n$ samples be represented as $\mathcal{D} = \{p_i,l_i\}_{i=1}^n$, where $p_i$ is the LiDAR scene consisting of 3D points that describe the scene, and $l_i$ consists of object labels such that each object in the scene is described with the label $ \{x,y,z,dx,dy,dz,\theta,c\} $ where $\{x,y,z\}$ denote the center of the bounding box, and $\{dx,dy,dz\}$ denote the displacement of the bounding box edges from the center to represent bounding box dimension, $\theta$ denotes the orientation, and $c$ denotes the class label.

Borrowing the notation from \cite{arjovsky2019invariant}, we consider a set of LiDAR point cloud datasets $\mathcal{D}_e$ consisting of $n_e$ labeled samples of $c$ object categories such that $ \mathcal{D}_e = \{p^e_i,l^e_i\}_{i=1}^{n_e}$, where $e \in \mathcal{E}_{tr}$ denotes the data acquisition environments of the training datasets. We assume the occurrence of the same object categories across training and evaluation datasets. We call the labeled datasets seen during training as source datasets which are samples from source domain distributions, and denote by $n_s = |\mathcal{E}_{tr}|$, the number of source domains available for training. The goal of DG is to train a network on these samples to perform well on unseen environments from unknown distributions (\ie, target domains).

\noindent\textbf{Region-level supervised contrastive learning.}
A LiDAR scene contains some regions that provide little information, due to the sparse nature of the point clouds. This makes global scene-level comparisons of samples from different distributions difficult beyond simplistic attributes such as point cloud density. We propose to focus on meaningful regions in the LiDAR scene, \ie, regions with objects of interest such as cars, bicycles, and pedestrians. Point cloud objects from the scene belonging to the same category tend to share geometric properties. A point cloud of a car is identifiable as a car even in the absence of visual attributes such as color and texture. Thus, we hope to learn a universal representation of a LiDAR ``Car'' object that is consistent across domains. We wish to train a network to learn region features that capture this domain invariant representation of each object category. From our multi-source dataset with samples captured in different environments, we have access to different data representations of the same object category. By pushing together objects from the same category and pulling apart objects from different categories, features across domains are aligned in a common embedding space.

\noindent\textbf{Feature selection.} 
Anchor-based networks provide region proposals for a given scene. These proposals consist of features corresponding to foreground as well as background regions. Given the ground-truth bounding box information, we can identify the most accurate proposals and use them for contrastive training. Consider a batch of input point clouds from different source domains $\{p_i^{e_j},l_i^{e_j}\}_{i=1}^b$ where $j$ indexes the source domains and $b$ is the batch size. Let the RPN head be denoted by $\phi_{rpn}$. After feature extraction and region pooling, $\phi_{rpn}$ generates a set of ROIs represented by features $f$ and predicted bounding boxes $B$ which we denote by 
$ \{\{f^{e_j}_{k},B^{e_j}_{k}\}_{k=1}^{n_{r}}\}_{j=1}^{n_s}$  for a given set of samples in a batch, where $e$ denotes the environment/domain of the sample, and $n_{r}$ and $n_s$ are the numbers of ROIs and domains, respectively.  
Many region proposals are redundant or belong to background areas, so we select the foreground ROIs closest to the nearest ground truth label, and sample $m$ background proposals. We now have a subset of  ROI features, ground truth class labels $c$, and domain labels from samples across different source distributions denoted by $ \{\{f^{e_j}_{k},c^{e_j}_{k}\}_{k=1}^{n_{r} + m}\}_{j=1}^{n_s}$.

\noindent\textbf{Supervised contrastive learning.}
We now have the region features of objects across samples in the batch, which originate from different domains. We enforce domain invariance in the learned features by  minimizing the distance between region features of objects from the same category and maximizing the distance between objects of different categories on ROIs prior to being passed to further layers. Consider the set of region features and corresponding category labels from scenes across different source distributions $\{\{f^{e_j}_{k},c^{e_j}_{k}\}_{k=1}^{n_{r}}\}_{j=1}^{n_s}$. To draw an analogy to contrastive learning frameworks like \cite{khosla2020supervised}, each region from each LiDAR scene is treated as a training sample, with the input batch consisting of $N$ regions (samples). Each sample has many positive pairs, decided by the number of samples sharing the same object category, and an arbitrary number of negative pairs. Unlike \cite{khosla2020supervised}, we do not have augmented views, since we obtain samples directly in the features space. We now show the way we perform contrastive learning which can handle multiple positives. 

To enforce domain invariance, distances between region features across LiDAR scenes are optimized by comparing region features from two source domains. 

\noindent\textbf{Computing feature similarity:}
A straightforward way of examining the relationship between features in a common embedding space is to directly compare the similarity measure between them. We use the cosine similarity for the unit-normalized feature vectors computed using $\text{sim} =  \langle f^{e_1} , f^{e_2} \rangle$.

\noindent\textbf{Optimizing feature similarity:} 
Inspired by the multimodal self-supervised contrastive learning method CLIP \cite{radford2021clip}, we utilize binary cross entropy (BCE) loss to optimize feature similarity. We modify the formulation to handle multiple positive pairs and compute the sum of BCE loss for sample pairs that are constructed for a particular class across all object categories, such that:
\begin{align}
\label{eq:l_bce}
\mathcal{L}_{con} = \sum\limits_{i=1}^{N_c} {\frac{1}{\left | R(i) \right |}}\sum\limits_{\substack{j \in R(i); h \in \textbf{R}\\ k,l \in D(i)}}-\omega_i \{q_iy_{j}\log(s) \nonumber +  (1-y_{j})\log(1-s)\},
\end{align}
where $s=\sigma(\text{sim}(f_{j}^{e_k},f_{h}^{e_l}))$, $\sigma(\cdot)$ denotes the softmax operation, $q_i$ is the weight value for positive samples computed by the ratio of number of negative and positive samples,  $\textbf{R}$ denotes the indices of all foreground regions and sampled background regions, and $y_j=\mathbb{1}(j=h)$ with $\mathbb{1}(\cdot)$ as the indicator function, $N_c$ denotes the number of classes, $R(i)$ and $D(i)$ denote the regions of interest and their respective domain labels. Optimizing these loss functions trains the feature extraction and region proposal branches to produce aligned feature representations for all regions in the batch that belong to the same object category, including those from different domains. These region features are then passed to a bounding box regression branch that performs localization, and a classification branch that categorizes each proposal.

\noindent\textbf{Losses for object detection:}
In addition to the contrastive loss functions, the network is supervised by the localization ($\mathcal{L}_{loc}$) and classification ($\mathcal{L}_{cls}$) losses present in their original training frameworks. The contrastive loss $\mathcal{L}_{con}$ is added to the existing RPN loss $\mathcal{L}_{rpn}$ and made to share a similar scale in value. The final loss $\mathcal{L}$ below is used to train the network: 
\begin{equation}
\mathcal{L} = \mathcal{L}_{loc} + \mathcal{L}_{cls} + \mathcal{L}_{rpn} + \mathcal{L}_{con}.
\end{equation}

\section{Experiments}

\subsection{Datasets}

In autonomous driving datasets, the environment of data capture is characterized by a variety of variables, including geographic location, type of LiDAR sensor, position of the sensor, and weather conditions during capture. These vary between datasets, providing a diverse set of training data with which to train a learning algorithm. For the purpose of DG, we seek diverse datasets that express different forms of distribution shift, \eg, from average size of vehicles to point cloud density. For our experiments, we choose four popular autonomous driving LiDAR-image datasets for 3D object detection: Lyft \cite{lyft}, KITTI \cite{KITTI}, Waymo Open Dataset \cite{waymo}, and nuScenes \cite{nuscenes2019}. In the supplement, we compare various properties of these datasets. We construct our domain shift scenarios to cover shifts from larger annotation-rich source datasets to smaller annotation-poor datasets as well as the reverse case. Due to their varying data formats, to train on multiple datasets at a time, we convert all the datasets to the format of KITTI for both training and evaluation.

\subsection{Experimental setup}
\label{sec:exp_set}
\noindent\textbf{Domain shift scenarios.}
To demonstrate our DG method, we explore cross-dataset distribution shift that covers a change in location, time of day, weather conditions, and rates of LiDAR return. We conduct transfer from multiple source datasets to a single target dataset at a time in the form $\{S_1,S_2\} \xrightarrow{} T$ where $S$ denotes the source dataset and $T$ denotes the target dataset. In particular, we conduct experiments with the settings:
\begin{enumerate}
[topsep=0pt,itemsep=-1ex,partopsep=1ex,parsep=1ex]
    \item $S_1 = \text{Waymo}, \; S_2 = \text{nuScenes}; \; T\in \{\text{Lyft},\text{KITTI}\}$
    \item $S_1 = \text{Lyft}, \; S_2 = \text{KITTI}; \; T\in \{\text{Waymo},\text{nuScenes}\}$
\end{enumerate}

This covers a broad variety of types of cross-dataset shift, including transferring to and from the smaller annotation-poor datasets such as KITTI and between datasets with dense and sparse point clouds. For baseline experiments, we also conduct single-source domain transfer experiments. A network trained on a particular source dataset setting may be evaluated on any target distribution.

\noindent\textbf{3D object detection networks:}
We demonstrate our multimodal domain generalization framework primarily on VoxelRCNN \cite{deng2021voxel}, which we choose due to its superior baseline object detection performance, the symmetry of the 3D feature extraction network to that of ResNet \cite{he2016deep}, and the existence of a region proposal network, which we require to leverage region features. We also demonstrate the pure-LiDAR approach on Part-$A^2$ for consistent comparison.

\noindent\textbf{LiDAR network architecture:}
The 3D feature extraction backbone consists of 4 stages of sequential sparse 3D convolutional blocks that produce feature maps of dimensions $16$, $32$, $64$, and $128$ with successively decreasing spatial dimensions. In terms of voxel features, set abstraction at each stage results in sparser voxels in scaled down ranges. This architecture may be observed in several 3D object detector networks such as \cite{shi2020pv,lang2019pointpillars,bhattacharyya2020deformable,deng2021voxel}.

\noindent\textbf{Image network architecture:}
We leverage a ResNet-50 backbone \cite{he2016deep} pre-trained for image segmentation on COCO \cite{lin2014microsoft} under DeepLabv3 \cite{chen2017deeplab} as the 2D image extraction backbone. This branch is trained end-to-end.

\begin{table*}[t]
\resizebox{\textwidth}{!}{%
\begin{tabular}{cccccccccccc}
\toprule
\multirow{2.5}{*}{Dataset}  & \multirow{2.5}{*}{Modality} & \multirow{2.5}{*}{Method}    
                          & \multicolumn{3}{c}{Car}                                              & \multicolumn{3}{c}{Pedestrian}                                       & \multicolumn{3}{c}{Cyclist}                      \\ \cmidrule(l){4-6} \cmidrule(l){7-9} \cmidrule(l){10-12}
                          &                           &                         & easy           & moderate           & \multicolumn{1}{c}{hard}           & easy           & moderate           & \multicolumn{1}{c}{hard}           & easy           & moderate           & hard           \\ \midrule
\multirow{4}{*}{KITTI}    & L                         & VoxelRCNN               & 92.36          & 82.74          & \multicolumn{1}{c}{80.04}          & 65.85          & \textbf{58.67} & \multicolumn{1}{c}{52.35}          & 88.05          & 69.53          & 64.92          \\
                          & L + I                     & VoxelRCNN - DeepFusion*  &    92.19            &       82.59         & \multicolumn{1}{c}{80.06}               &     60.56           &    52.82            & \multicolumn{1}{c}{46.63}               &       84.79         &      62.70          &    58.61            \\
                          & L + I                     & VoxelRCNN - FocalsConv     & \textbf{92.55} & 82.92          & \multicolumn{1}{c}{80.34}          & 61.01          & 53.28          & \multicolumn{1}{c}{48.02}          & \textbf{89.17} & \textbf{70.26} & 65.67          \\
                          & L + I                     & VoxelRCNN - MSFusion (ours)          & 92.22          & \textbf{83.32} & \multicolumn{1}{c}{\textbf{82.45}} & \textbf{66.19} & 58.41          & \multicolumn{1}{c}{\textbf{52.63}} & 89.10           & 70.23          & \textbf{66.34} \\ \midrule
\multirow{4}{*}{Waymo}    & L                         & VoxelRCNN               & 76.19          & 74.59          & \multicolumn{1}{c}{72.17}          & 57.89          & 58.25          & \multicolumn{1}{c}{57.88}          & 65.79          & 66.36          & 66.26          \\
                          & L + I                     & VoxelRCNN - DeepFusion*  &    \textbf{76.41}            &     74.82          & \multicolumn{1}{c}{72.51}               &     56.67           &     56.40           & \multicolumn{1}{c}{56.05}               &     64.97           &     65.80           &       63.29         \\
                          & L + I                     & VoxelRCNN - FocalsConv     &        76.40	       &     \textbf{76.07}	            & \multicolumn{1}{c}{\textbf{72.81}}               &     57.18	          &        57.62	        & \multicolumn{1}{c}{\textbf{57.93} }               &     68.39	           &       66.60         &        66.06        \\
                          & L + I                     & VoxelRCNN - MSFusion (ours)          & 76.17          & 74.74          & \multicolumn{1}{c}{72.38}          & \textbf{58.49}          & \textbf{58.31}          & \multicolumn{1}{c}{57.66}          & \textbf{73.89}          & \textbf{71.97}          & \textbf{68.04}          \\ \midrule
\multirow{4}{*}{nuScenes} & L                         & VoxelRCNN               & 22.05          & 18.35          & \multicolumn{1}{c}{17.30}           & 8.99           & 7.98           & \multicolumn{1}{c}{7.52}           & 8.48           & 7.42           & 6.69           \\
                          & L + I                     & VoxelRCNN - DeepFusion*  &    24.44            &      20.52        & \multicolumn{1}{c}{19.14}               &      \textbf{13.99}          &       \textbf{12.59}         & \multicolumn{1}{c}{\textbf{11.95}}               &     9.62           &      8.97          &      8.01          \\
                          & L + I                     & VoxelRCNN - FocalsConv     &      20.74          &       17.46         & \multicolumn{1}{c}{16.28}               &        8.59        &          7.80      & \multicolumn{1}{c}{7.49}               &         11.55       &      10.69          &      9.47          \\
                          & L + I                     & VoxelRCNN - MSFusion (ours)          & \textbf{25.69}         & \textbf{21.50}           & \multicolumn{1}{c}{\textbf{20.41}}          & 9.45           & 8.58           & \multicolumn{1}{c}{8.55}           & \textbf{12.28}          & \textbf{11.44}          & \textbf{10.01}          \\ \midrule
\multirow{4}{*}{Lyft}     & L                         & VoxelRCNN               & 85.91          & 75.44          & \multicolumn{1}{c}{71.57}          & 45.77          & 33.74          & \multicolumn{1}{c}{32.75}          & 73.95          & 54.93          & 51.23          \\
                          & L + I                     & VoxelRCNN - DeepFusion*  &     88.67           &        78.64        & \multicolumn{1}{c}{76.68}               &    53.53            &       40.97         & \multicolumn{1}{c}{39.88}               &      77.55          &       60.67         &    56.77            \\
                          & L + I                     & VoxelRCNN - FocalsConv     & 88.99          & 79.26          & \multicolumn{1}{c}{77.29}          & \textbf{56.59} & 40.86          & \multicolumn{1}{c}{40.10}           & \textbf{78.91} & \textbf{62.87} & 58.84          \\
                          & L + I                     & VoxelRCNN - MSFusion (ours)          & \textbf{89.11} & \textbf{79.86} & \multicolumn{1}{c}{\textbf{78.00}}    & 55.53          & \textbf{43.19} & \multicolumn{1}{c}{\textbf{42.27}} & 78.56          & 62.51          & \textbf{59.15} \\ \bottomrule
\end{tabular}}
\caption{3D average precision (AP) results of VoxelRCNN \cite{deng2021voxel} trained on KITTI \cite{KITTI}, Waymo \cite{waymo}, nuScenes \cite{nuscenes2019}, and Lyft \cite{lyft}. Compares performance of training the network with just LiDAR (L) or LiDAR and image (L+I) using DeepFusion \cite{li2022deepfusion} and FocalsConv \cite{chen2022focal} with the proposed MSFusion approach. $*$ indicates the method is re-implemented by us. This table shows that (a) fusion of LiDAR and RGB inputs leads to improved performance, and (b) the proposed fusion method outperforms the SOTA methods.}
\label{tab:fusion}
\vspace{-0.16in}
\end{table*}

\noindent\textbf{Baselines and oracle:}
We compare our multi-source contrastive learning-based DG framework against single-source and multi-source direct transfer baselines, in which a source-trained model is directly evaluated on the target datasets. That is, in ``direct transfer,'' the object detection networks are trained only with the standard classification and bounding box regression loss functions, without contrastive loss. We also compare our method against the single-source DG work 3D-VField for the backbone network Part-$A^2$ which uses adversarial data augmentation. As the implementation of 3D-VField has not been made public, we compare our results against those reported in the paper. We also provide the ``oracle'' results for both LiDAR and LiDAR-image detectors, \ie, when both source and target domains are the same, and trained using only classification and bounding box regression losses. This can be observed in Table \ref{tab:fusion}.

\noindent\textbf{Evaluation metric:}
We evaluate the object detection networks in average precision (AP) in the KITTI format. We report performance on the object categories of ``Car,'' ``Pedestrian,'' and ``Cyclist'' with respective 3D IoU thresholds of $0.7$, $0.5$, $0.5$ for the KITTI, Lyft, and Waymo datasets and $0.5$, $0.25$, $0.25$ for the nuScenes dataset, respectively. For a fair comparison with the reported numbers of \cite{lehner20223d}, we include evaluation results on Waymo with the 3D IoU threshold $0.5$ for the ``moderate'' category in Table \ref{tab:pa2}. Each object is further categorised in terms of difficulty as ``easy,'' ``moderate,'' or ``hard'' based on the level of occlusion, truncation, and distance from the camera. We follow \cite{KITTI} for this evaluation convention.

\noindent\textbf{Implementation details:}
We train the object detection networks in our supervised contrastive learning framework on an equal number of samples from each source dataset, and ensure that an equal number of scenes from each dataset make up an input batch. Since the quality of the selected region features depends on the proposal network, we pre-train the network for 30 epochs on the source datasets without contrastive loss before training for a further 30 epochs with the contrastive loss. We use the standard data augmentation methods for object detectors such as global scaling, rotation, and ground-truth sampling. We train the networks with a batch size of 32, with the Adam optimizer \cite{kingma2014adam} and a cyclic learning rate scheduler. The initial learning rate is $0.01$ with a weight decay of $0.01$.  For the implementation of the object detectors, we follow the codebase OpenPCDet \cite{openpcdet2020}. We train each model on four 48GB NVIDIA A40 GPUs.

\begin{table*}[]
\centering
\resizebox{\textwidth}{!}{%
\begin{tabular}{cccccccccccc}
\toprule
\multirow{2.5}{*}{Train / Test}             & \multirow{2.5}{*}{Modality} & \multirow{2.5}{*}{Method} & \multicolumn{3}{c}{Car}                   & \multicolumn{3}{c}{Pedestrian}            & \multicolumn{3}{c}{Cyclist}    \\ \cmidrule(l){4-6} \cmidrule(l){7-9} \cmidrule(l){10-12} 
                                        &                           &                         & easy  & moderate  & \multicolumn{1}{c}{hard}  & easy  & moderate  & \multicolumn{1}{c}{hard}  & easy  & moderate  & hard           \\ \midrule
\multirow{2}{*}{Waymo / KITTI}            & L                         & DT                      & 16.12  & 14.95 & \multicolumn{1}{c}{14.43} & 60.39 & 53.01 & \multicolumn{1}{c}{47.89} & 67.51 & 57.40 & 51.96          \\
                                        & L + I                     & DT                   & 6.33  & 5.98  & \multicolumn{1}{c}{5.48}  & 57.26 & 50.69 & \multicolumn{1}{c}{45.93}    & 67.75 & \textbf{62.91} & \textbf{58.11}          \\ \cdashlinelr{1-12}
\multirow{2}{*}{nuScenes / KITTI}         & L                         & DT                      & 3.05  & 2.89  & \multicolumn{1}{c}{2.58}  & 34.03 & 28.92 & \multicolumn{1}{c}{25.93} & 7.05 & 4.60  & 4.50          \\
                                        & L + I                     & DT                   & 12.51  & 8.92  & \multicolumn{1}{c}{7.41}  & 31.31 & 25.87 & \multicolumn{1}{c}{23.52} & 28.56 & 16.45 & 15.84 \\ \cdashlinelr{1-12}
\multirow{2}{*}{Waymo + nuScenes / KITTI} & L                         & \method~                 & \textbf{37.90}  & \textbf{35.94} & \multicolumn{1}{c}{\textbf{36.77}} & 58.92 & 51.81 & \multicolumn{1}{c}{45.94} & \textbf{70.43} & 57.91 & 54.35          \\
                                        & L + I                       & \method~              & 37.72 & 35.22 & \multicolumn{1}{c}{36.37} & \textbf{65.07} & \textbf{57.94} & \multicolumn{1}{c}{\textbf{51.84}} & 68.71 & 55.33 & 52.35          \\ \midrule
\multirow{2}{*}{Waymo / Lyft}             & L                         & DT                      & 60.08 & 41.99 & \multicolumn{1}{c}{41.62} & 38.43 & 23.80 & \multicolumn{1}{c}{23.55} & 41.95 & 22.17 & 20.08          \\
                                        & L + I                     & DT                   & 59.33 & 42.31 & \multicolumn{1}{c}{41.95} & 37.25 & 21.88 & \multicolumn{1}{c}{21.86} & \textbf{46.71} & 25.99 & 23.39          \\ \cdashlinelr{1-12}
\multirow{2}{*}{nuScenes / Lyft}          & L                         & DT                      & 22.67 & 13.77 & \multicolumn{1}{c}{14.12} & 7.55  & 3.98  & \multicolumn{1}{c}{4.23}  & 7.15  & 3.84  & 3.21           \\
                                        & L + I                     & DT                   & 36.15 & 23.74 & \multicolumn{1}{c}{23.96} & 7.07 & 4.67  & \multicolumn{1}{c}{4.51}   & 11.07 & 5.80 & 5.08          \\ \cdashlinelr{1-12}
\multirow{2}{*}{Waymo + nuScenes / Lyft}  & L                         & \method~                 & 56.85 & 39.66 & \multicolumn{1}{c}{38.79} & 35.57 & 22.74 & \multicolumn{1}{c}{22.47} & 45.86 & \textbf{26.96} & \textbf{26.11}          \\
                                        & L + I                       & \method~              & \textbf{63.87} & \textbf{44.18} & \multicolumn{1}{c}{\textbf{43.02}} & \textbf{39.38} & \textbf{25.22} & \multicolumn{1}{c}{\textbf{24.87}} & 37.82 & 21.64 & 20.46          \\ \midrule
\multirow{2}{*}{KITTI / Waymo}            & L                         & DT                      &  14.10 & 14.53 & \multicolumn{1}{c}{13.89} & 32.56 & 33.16 & \multicolumn{1}{c}{31.32} & 30.46 & 27.46 &    24.51       \\
                                        & L + I                     & DT                   & 18.62  &  18.63 & \multicolumn{1}{c}{17.67}  & 22.60 & 23.48 & \multicolumn{1}{c}{23.66}    &  36.35& 35.44 &    32.46       \\ \cdashlinelr{1-12}
                            
\multirow{2}{*}{Lyft / Waymo}         & L                         & DT                      &  46.78 & 45.21  & \multicolumn{1}{c}{41.53}  & 27.99 & 28.69 & \multicolumn{1}{c}{28.85} & 21.34 & 22.81  & 22.47           \\

                                        & L + I                     & DT                   &  \textbf{54.98} &  \textbf{54.93} & \multicolumn{1}{c}{\textbf{51.73}}  & 37.06 & 38.54 & \multicolumn{1}{c}{38.24} & 31.29 & 31.41 & 32.16 \\ \cdashlinelr{1-12}
\multirow{2}{*}{KITTI + Lyft / Waymo} & L                         & \method~                 &  28.13 & 28.06 & \multicolumn{1}{c}{25.18} & 43.12 & 43.43 & \multicolumn{1}{c}{42.51} & 47.17 & 43.53 &    41.33       \\ 
                                        & L + I                       & \method~              & 48.13& 46.57 & \multicolumn{1}{c}{43.65} & \textbf{47.62} & \textbf{46.93} & \multicolumn{1}{c}{\textbf{45.59}} &  \textbf{48.06}& \textbf{49.03} &   \textbf{46.05}        \\ \midrule
\multirow{2}{*}{KITTI / nuScenes}             & L                         & DT                      & 24.75& 20.70 & \multicolumn{1}{c}{19.12} & 9.26 & 8.09 & \multicolumn{1}{c}{8.09} & 3.45 & 3.67 & 3.55          \\
                                        & L + I                     & DT                   & 28.23 & 23.15 & \multicolumn{1}{c}{21.56} & \textbf{9.63} & \textbf{8.21} & \multicolumn{1}{c}{\textbf{8.17}} & 4.30 & 4.62 &    3.90      \\ \cdashlinelr{1-12}
\multirow{2}{*}{Lyft / nuScenes}          & L                         & DT                      & 21.36 & 17.35 & \multicolumn{1}{c}{16.21} & 6.33  & 6.42 & \multicolumn{1}{c}{6.53}  &  4.36 &  5.25 &   4.74         \\
                                        & L + I                     & DT                   & 24.25 & 19.95 & \multicolumn{1}{c}{18.63} & 9.03 &  7.57 & \multicolumn{1}{c}{7.44}   & 6.55 & 5.89 &   5.30        \\ \cdashlinelr{1-12}
\multirow{2}{*}{KITTI + Lyft / nuScenes}  & L                         & \method~                 & 25.02 & 20.60 & \multicolumn{1}{c}{19.41} & 5.63 & 5.49 & \multicolumn{1}{c}{4.32} & 3.44 & 3.62 &      2.96     \\
                                        & L + I                       & \method~              & \textbf{31.06} & \textbf{25.41} & \multicolumn{1}{c}{\textbf{24.09}} & 4.82 & 4.79 & \multicolumn{1}{c}{4.82} & \textbf{7.10} & \textbf{6.80} &   \textbf{6.07}        \\ \bottomrule
\end{tabular}}
\caption{3D average precision (AP) of unseen domain transfer to the Waymo, nuScenes, KITTI and Lyft datasets. DT = direct transfer, L=LiDAR, I=Image. We show that multi-source training of multimodal detectors under \method~'s contrastive learning framework results in improved robustness to unseen domains.}
\vspace{-0.2in}
\label{tab:wnkl}
\end{table*}

\section{Results and discussion}
We present and discuss the results of the MSFusion method on each LiDAR-image dataset as well as the full \method~ framework evaluated on various domain shifts. We point to the supplementary material for qualitative results.

\noindent\textbf{LiDAR-RGB fusion.}
In Table \ref{tab:fusion} we compare the proposed MSFusion method for fusing LiDAR and RGB with the SOTA convolution-based multimodal object detectors that use single view images, DeepFusion \cite{li2022deepfusion} and FocalsConv \cite{chen2022focal}, as well as the pure-LiDAR baseline for VoxelRCNN. In this setting, the training and testing domains are the same, and we convert each dataset to KITTI FOV format with standard splits. We observe consistent improvement from the pure-LiDAR baseline, particularly in the ``hard'' category of objects, which are characterized by high occlusion, truncation, and larger distances from the camera. We show that multi-stage fusion of LiDAR-RGB features outperforms or matches the SOTA in most cases. In cases where the proposed method does not perform the best, it comes second with only a small margin. 

\begin{wraptable}{r}{6.5cm}
\centering
\resizebox{\linewidth}{!}{%
\begin{tabular}{ccccc}
\toprule
\multirow{2.5}{*}{Train / Test}    & \multirow{2.5}{*}{Method} & \multicolumn{3}{c}{AP}       \\ \cmidrule(l){3-5} 
                               &                         & Car   & Pedestrian & Cyclist \\ \midrule
\multirow{2}{*}{KITTI / Waymo} & DT                      & 47.64 & 26.45      & 53.83   \\
                               & 3D-Vfield \cite{lehner20223d}               & 56.80  & -          & -       \\ \cdashlinelr{1-5}
Lyft / Waymo                   & DT                      & 56.97 & 22.44      & 48.42   \\ \cdashlinelr{1-5}
\multirow{2}{*}{KITTI + Lyft / Waymo}    & DT                      & 65.34 & 32.41      & \textbf{67.78}  \\
                               & \method~                 & \textbf{66.59} & \textbf{34.68}      & 66.28   \\ \bottomrule
\end{tabular}
}
\caption{3D detection results for IoU thresholds $\{0.5,0.5,0.25\}$ on LiDAR cross-domain experiments on Part-A$^2$ \cite{shi2021parta2}. DT = direct transfer.}
\label{tab:pa2}
\vspace{-0.1in}
\end{wraptable}

\noindent\textbf{Detection on unseen domains.}
We demonstrate our multimodal framework for four domain shift scenarios. We compare our method against direct transfer baselines and that of 3D-Vfield \cite{lehner20223d}. By direct transfer, we mean that the network is trained on the source dataset and evaluated on the target dataset with no adaptation or changes to training. In Table \ref{tab:wnkl} we show the performance of VoxelRCNN \cite{deng2021voxel} trained under the domain shift listed in Sec. \ref{sec:exp_set}. We point out observations that are two-fold. Firstly, introducing image information improves domain transfer performance in most cases. Particularly for the ``Cyclist'' category, we observe a large performance improvement when evaluating on KITTI and Lyft. This supports the suggestion that multimodal detectors are more robust to unseen distributions. Secondly, training the object detector under the multi-source contrastive framework also aids in improved robustness to out-of-distribution samples. Compared to single source direct transfer cases, we observe a significant performance improvement in most cases when the network is trained under \method~, and this trend continues when applied to the multimodal detector. This indicates that encouraging domain invariance in a diverse training set can aid generalization. In Table \ref{tab:pa2}, we compare our LiDAR-only contrastive learning framework on the detector Part-A$^2$ with the domain generalization work 3D-Vfield \cite{lehner20223d}. Since the code is not publicly available, we compare against the reported domain scenario and single category. We outperform \cite{lehner20223d}, but point out that their learnable data augmentations may be incorporated into our framework.

\begin{figure}
    \centering
    \includegraphics[width=\linewidth]{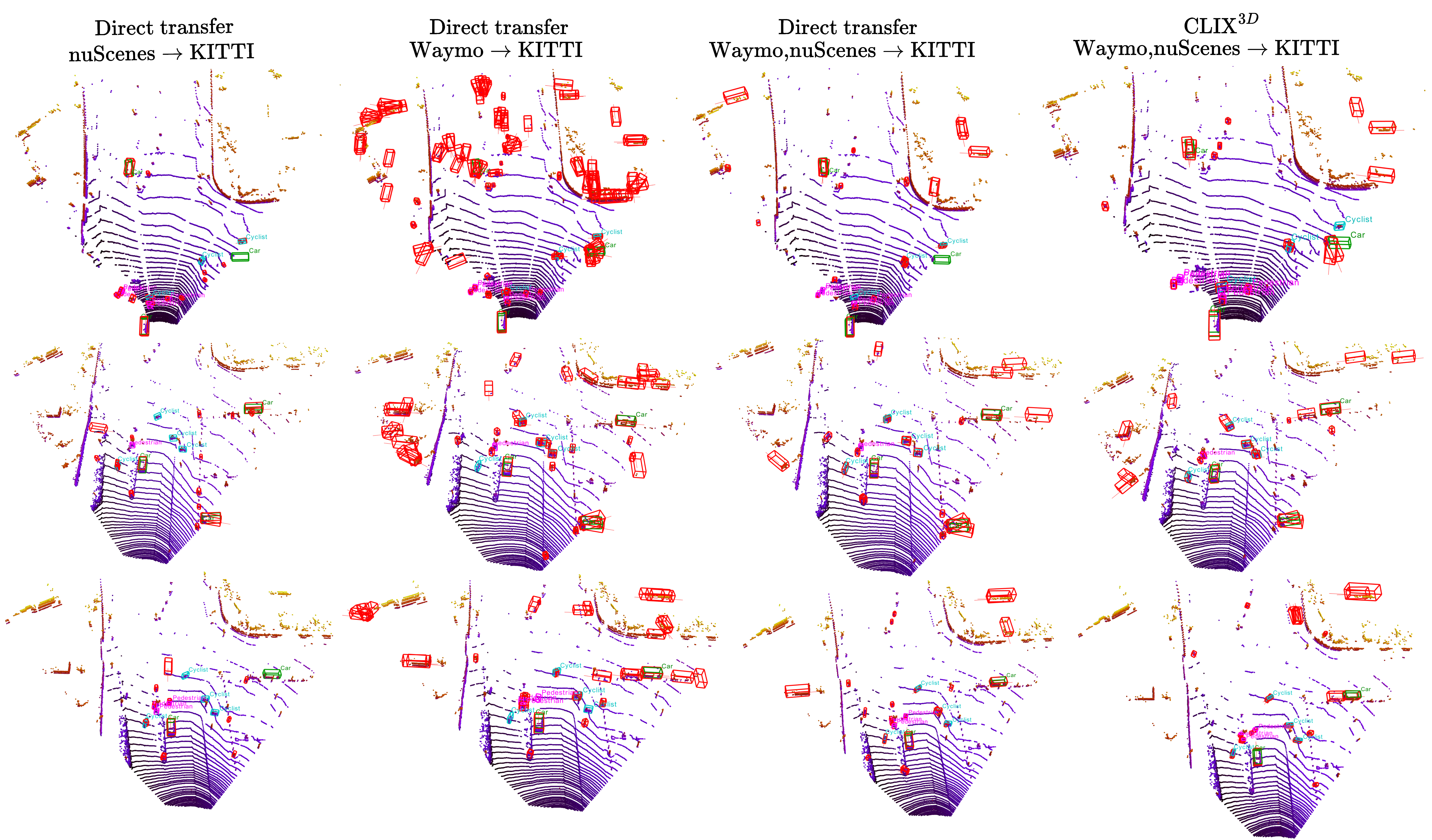}
    \caption{A qualitative comparison of the detection results of Part-$A^2$ trained for the domain shift scenario $\text{Waymo, nuScenes}\rightarrow \text{KITTI}$. Ground truth bounding boxes for the ``Car'' category are in \green{green}, in \pink{magenta} for the ``Pedestrian'' category, and in \cyan{cyan} for the ``Cyclist'' category. Predictions are in \red{red}. (Best viewed zoomed in and in color). }
    \label{fig:qual}
\end{figure}

\noindent\textbf{Qualitative evaluation of 3D object detection.}
We provide a qualitative comparison of the detection results of our proposed method \method~ against those of the single and multi-source direct transfer (DT) baselines for the Part-$A^2$ network for the domain shift scenario of $\text{Waymo, nuScenes}\rightarrow \text{KITTI}$.  This comparison is shown in Fig. \ref{fig:qual}, where the columns correspond to the results from each method, while each row corresponds to the samples from the KITTI validation dataset. We visualize the bounding boxes that are predicted with a confidence score greater that $0.3$. Our method \method~ addresses the problem of missed detections (false negatives) as well as superfluous predictions (false positive) faced by the baseline approaches that affect the precision score. The DT $\text{Waymo}\rightarrow \text{KITTI}$ method in particular predicts numerous false positives with high confidence.  The DT $\text{nuScenes}\rightarrow \text{KITTI}$ model does not suffer from false positives, but fails to predict most instance of the ``Cyclist'' class (see column 1, rows 2 and 3). Multi-source DT (column 3) addresses some of these problems but still fails to detect some instance of ``Car'' and ``Pedestrian.'' Column 4 shows the qualitative improvement our method, which predicts more instance of ``Pedestrian'' with fewer false positives of the ``Car'' category.

\noindent\textbf{Ablation study.}
We examine each component of the multimodal multi-source contrastive learning framework and demonstrate empirically the role they play in training a generalizable object detection network. In Table \ref{tab:ab1}, we compare the transfer performance to the Lyft dataset under different modules of the proposed framework. We find that adding the contrastive objective aids in multi-source training.

\begin{table}[t]
\centering
\resizebox{.65\linewidth}{!}{%
\begin{tabular}{cccccc}
\toprule
\multirow{2}{*}{Multi-source} & \multirow{2}{*}{Multimodal} & \multirow{2}{*}{Contrastive loss} & \multicolumn{3}{c}{Car} \\ \cline{4-6} 
                              &                              &                                   & easy   & moderate   & hard   \\ \hline

$\ccheck$            & \ccross          & \ccross               &  59.11  &      40.53     &   39.53   \\
\ccross            & $\ccheck$           & \ccross                &  59.95  &     41.39      &   40.55 \\
$\ccheck$           & $\ccheck$           & \ccross                & 62.07  &     43.08     & 41.79  \\

$\ccheck$            & $\ccheck$           & $\ccheck$                &  \textbf{63.87}  &      \textbf{44.18}      &    \textbf{43.02}     \\  \bottomrule
\end{tabular}}
\caption{Ablation study for Waymo+nuScenes transfer to the Lyft dataset with 3D precision values for the ``Car'' category. Single source experiments indicate source is Waymo.}
\vspace{-0.15in}
\label{tab:ab1}
\end{table}

\noindent\textbf{Limitations.}
We observe some inconsistent performance when training with multiple sources. In the case of transferring to the Waymo dataset, we observe that the multi-source trained model performs worse than Lyft/Waymo direct in the ``Car'' category, but outperforms all single-source networks in the ``Pedestrian'' and ``Cyclist'' categories. This could be due to the fact that the average size of cars in the Lyft dataset is closer than KITTI to that of the Waymo dataset. We believe this problem can be mitigated by training with more diverse source data, and point out that in the DG setting, \emph{it is not possible to choose a single source that is closer in distribution to the target data}, in which case the multi-source training that performs better on average is preferred. We also observe that the ``Pedestrian'' and ``Cyclist'' categories of the nuScenes dataset are difficult to transfer to, resulting in relatively lower precision values. This is due to the sparse nature of the dataset, which we hope to explicitly address in future work.

\section{Conclusion}
For the problem of unseen domain shift in 3D object detection, we show that MSFusion, our proposed LiDAR+RGB fusion method, outperforms prior methods and that multimodal fusion improves robustness when the target domains are unseen during training. We further improve generalization by proposing a multi-source training framework \method~ with simple yet effective region-level contrastive learning, which promotes invariance among features belonging to the same class across domains and pushes features from different classes apart. \method~ beats state-of-the-art methods in most cases under many distribution shifts for multiple object classes and datasets.
\bibliographystyle{splncs04}
\bibliography{egbib}

\newpage

\section{Supplementary Material}

\begin{table*}
    \centering
    \resizebox{\linewidth}{!}{%
    \begin{tabular}{@{}c@{}c@{\hspace{2mm}}c@{\hspace{2mm}}c@{\hspace{2mm}}c@{}}
    \toprule
     &KITTI &Waymo &nuScenes &Lyft\\
    \midrule
    LiDAR sensor &Velodyne HDL-64 &1$\times$360$^{\circ}$, 4$\times$HoneyComb & Velodyne HDL-32 & 1$\times$64-beam, 2$\times$40-beam\\
    Point cloud size &100K &150K &70K &40K\\
    Point cloud range &[$0, -40, -3, 70.4, 40, 1$] & [$-75.2, -75.2, -2, 75.2, 75.2, 4$]&[$-51.2, -51.2, -5.0, 51.2, 51.2, 3.0$] &[$-80.0, -80.0, -5.0, 80.0, 80.0, 3.0$]\\
    LiDAR height &1.73 &3.33 &1.8 &1.45\\
     ``Car'' anchor   &[$3.90, 1.60, 1.56$] &[$4.70, 2.10, 1.70$] &[$4.63, 1.97, 1.74$] &[$4.75, 1.92, 1.71$]\\
     ``Cyclist'' anchor   &[$1.76, 0.60, 1.73$] &[$1.78, 0.84, 1.78$] &[$1.70, 0.60, 1.28$] & [$1.76, 0.63, 1.44$]\\
    ``Pedestrian'' anchor   &[$0.80, 0.60, 1.73$] &[$0.91, 0.86, 1.73$] &[$0.73, 0.67, 1.77$] &[$0.80, 0.76, 1.76$]\\
    $\#$ Annotated 3D bounding box &200K &12M &1.4M &1.3M\\
    Location of capture  &Germany &USA &USA, Singapore &USA\\
    Weather conditions & sunny &variety &variety &variety \\
    \bottomrule
    \end{tabular}}
    \caption{Comparison between the autonomous driving datasets used in our experiments. All distances are in meters.}
    \vspace{-3.7mm}
    \label{tab:datasets}
\end{table*}

\subsection{Datasets for experiments}

For our experiments, we choose four popular autonomous driving LiDAR-image datasets for 3D object detection: Lyft \cite{lyft}, KITTI \cite{KITTI}, Waymo Open Dataset \cite{waymo}, and nuScenes \cite{nuscenes2019}. In Table \ref{tab:datasets}, we compare various properties of these datasets. This includes the conditions of data capture, such as sensor specifications, location, and weather as well as the properties of the data itself such as the size of the scene and  the average dimensions of objects.

\subsection{Additional implementation details}
\noindent\textbf{Evaluation metrics} 
We report the 3D mean average precision of the ``Car,'' ``Pedestrian,'' and ``Cyclist'' categories at the medium difficulty, following the KITTI evaluation metric \cite{KITTI}. Since all networks are converted to the uniform format of the KITTI dataset, we use this same evaluation metric across all datasets, and consider only the image field-of-view for all lidar scenes. In the case of Part-$A^2$ evaluation on the Waymo \cite{waymo} dataset, we report performance at 3D IoU thresholds $0.5,0.25,0.25$ for the ``Car,'' ``Pedestrian,'' and ``Cyclist'' categories respectively. This is done to perform a fair comparison with 3D-Vfield \cite{lehner20223d}, which uses the same metric specification, and to be consistent for model selection.  

When performing domain transfer to the Waymo dataset, we lower the target point clouds and ground truth bounding boxes  by $1.6$m to align them with the ground planes of the source datasets of Lyft and KITTI. This is done during the evaluation step only, and is consistent with the procedure followed by Lehner \etal \cite{lehner20223d} in 3D-Vfield.

\end{document}